\documentclass[a4paper,fleqn]{cas-sc}

\usepackage{hyperref}
\usepackage[numbers]{natbib}
\usepackage{microtype}

\def\tsc#1{\csdef{#1}{\textsc{\lowercase{#1}}\xspace}}
\tsc{WGM}
\tsc{QE}

\begin{document}
\let\WriteBookmarks\relax
\def\floatpagepagefraction{1}
\def\textpagefraction{.001}
\shorttitle{}    

\shortauthors{}  
\title [mode = title]{A Lightweight Group Multiscale Bidirectional Interactive Network for Real-Time Steel Surface Defect Detection}  
\author[university1]{Yong Zhang}

\affiliation[university1]{
organization= {School of Computer Science, Chongqing University},
addressline= {No. 55, South University Road, High-tech Zone},
city={Chongqing},
postcode={401331},
state={Chongqing},
country={China}
}

\author[university2]{Cunjian Chen}
\author[university2]{Qiang Gao}
\author[university1]{Yi Wang}
\author[university1]{Bin Fang}[orcid=0000-0003-1955-6626] \corref{cor1} %

\affiliation[university2]{organization={Department of Data Science and AI, Monash University},
            addressline={Clayton Campus, 21 Chancellor's Walk, Clayton}, 
            city={Melbourne},
            postcode={3800}, 
            state={Victoria},
            country={Australia}}

\cortext[cor1]{Corresponding author} 
\ead{fb@cqu.edu.cn}

\shortauthors{Yong Zhang et~al.}
\begin{abstract}
Real-time surface defect detection is critical for maintaining product quality and production efficiency in the steel manufacturing industry. Despite promising accuracy, existing deep learning methods often suffer from high computational complexity and slow inference speeds, which limit their deployment in resource-constrained industrial environments. Recent lightweight approaches adopt multibranch architectures based on depthwise separable convolution (DSConv) to capture multiscale contextual information. However, these methods often suffer from increased computational overhead and lack effective cross-scale feature interaction, limiting their ability to fully leverage multiscale representations. To address these challenges, we propose GMBINet, a lightweight framework that enhances multiscale feature extraction and interaction through novel Group Multiscale Bidirectional Interactive (GMBI) modules. The GMBI employs a group-wise strategy for multiscale feature extraction, ensuring a consistent computational cost regardless of the number of scales. It further integrates a Bidirectional Progressive Feature Interactor (BPFI) and a parameter-free Element-Wise Multiplication-Summation (EWMS) operation to enhance cross-scale interaction without introducing additional computational overhead. Experimental results show that our method achieves a better trade-off between detection accuracy and computational efficiency than existing approaches across multiple steel surface defect datasets. The dataset and code are publicly available at: \url{https://github.com/zhangyongcode/GMBINet}.

\end{abstract}

\maketitle

\section{Introduction}\label{tab:introduction}

Steel is a fundamental critical material in modern industrial sectors, where surface quality directly affects the reliability and aesthetic quality of downstream products~\cite{shen2024minet}. While manual inspection is still employed in some production lines, it is labor-intensive, subjective, and inconsistent, making it unsuitable for the high-precision and real-time demands of modern large-scale production environments~\cite{song2020edrnet}. Traditional automated methods relying on handcrafted features have improved inspection efficiency to some extent \cite{zhu2014saliency, yu2010cluster}.  However, these methods suffer from limited feature representation capacity, leading to two critical drawbacks: poor robustness to environmental noise and inadequate generalization across defect scales \cite{yu2025lightweight}. In contrast, deep learning (DL) techniques, driven by their powerful feature extraction capabilities, have rapidly advanced the field of surface defect detection and offer broad potential for industrial applications \cite{ameri2024systematic, tulbure2022review, tang2023review}. 

Current DL-based methods for surface defect detection can be broadly categorized into two categories: performance-oriented and lightweight-oriented methods. Performance-oriented models (e.g., EDRNet \cite{song2020edrnet}, DACNet \cite{zhou2021dense}, EMINet \cite{zhou2021edge}) typically employ complex backbones (e.g., VGG \cite{VGG}, ResNet \cite{he2016deep}, Transformer \cite{liu2021swin}) with strong multiscale semantic extraction capabilities, enabling high detection accuracy. However, their substantial computational demands and inference latency pose significant challenges for real-time deployment, particularly on resource-constrained industrial devices \cite{shen2024minet}. To address these limitations, lightweight-oriented methods leverage cost-efficient backbones (e.g., MobileNet \cite{howard2017mobilenets}, ShuffleNet \cite{ma2018shufflenet}, GhostNet \cite{han2020ghostnet}) to reduce both parameter count and computational overhead. Despite their efficiency, these models often compromise accuracy,  especially when handling surface defects with significant scale variations \cite{shen2024minet}. To enhance scale awareness within lightweight networks and improve detection performance, recent efforts have introduced multibranch architectures based on depthwise separable convolution (DSConv), where each branch is designed to capture features at a distinct receptive field \cite{shen2024minet, liu2021samnet}. However, such designs introduce two significant limitations: (1)~\textbf{linearly increasing complexity}, as each branch processes the full input tensor, resulting in elevated memory consumption and latency with an increasing number of branches, contradicting the goal of lightweight design; and (2)~\textbf{limited cross-scale interaction}, as isolated branches hinder effective cross-scale communication, leading to redundant representations.

To overcome these limitations, we propose the Group Multiscale Bidirectional Interactive (GMBI) module, a novel lightweight structure that extends DSConv to support efficient multiscale feature extraction and interaction. Specifically, GMBI partitions the input tensor into channel-wise groups, with each group responsible for extracting features at a distinct scale. This design maintains channel capacity while reducing overall complexity. To facilitate cross-scale feature interaction, we introduce the Bidirectional Progressive Feature Interactor (BPFI), which enables hierarchical feature refinement through forward and backward paths. In the forward guidance path, smaller-scale features guide larger-scale representation learning through hierarchical priors, while in the backward path, larger-scale features refine spatial detail at lower levels via attention cues. Additionally, we design a lightweight, parameter-free fusion operation called Element-Wise Multiplication-Summation (EWMS) that emphasizes salient regions via element-wise multiplication and integrates complementary context through summation, enhancing discriminative representation at negligible computational cost.

\begin{figure}
  \centering
    \includegraphics[width=0.85\linewidth]{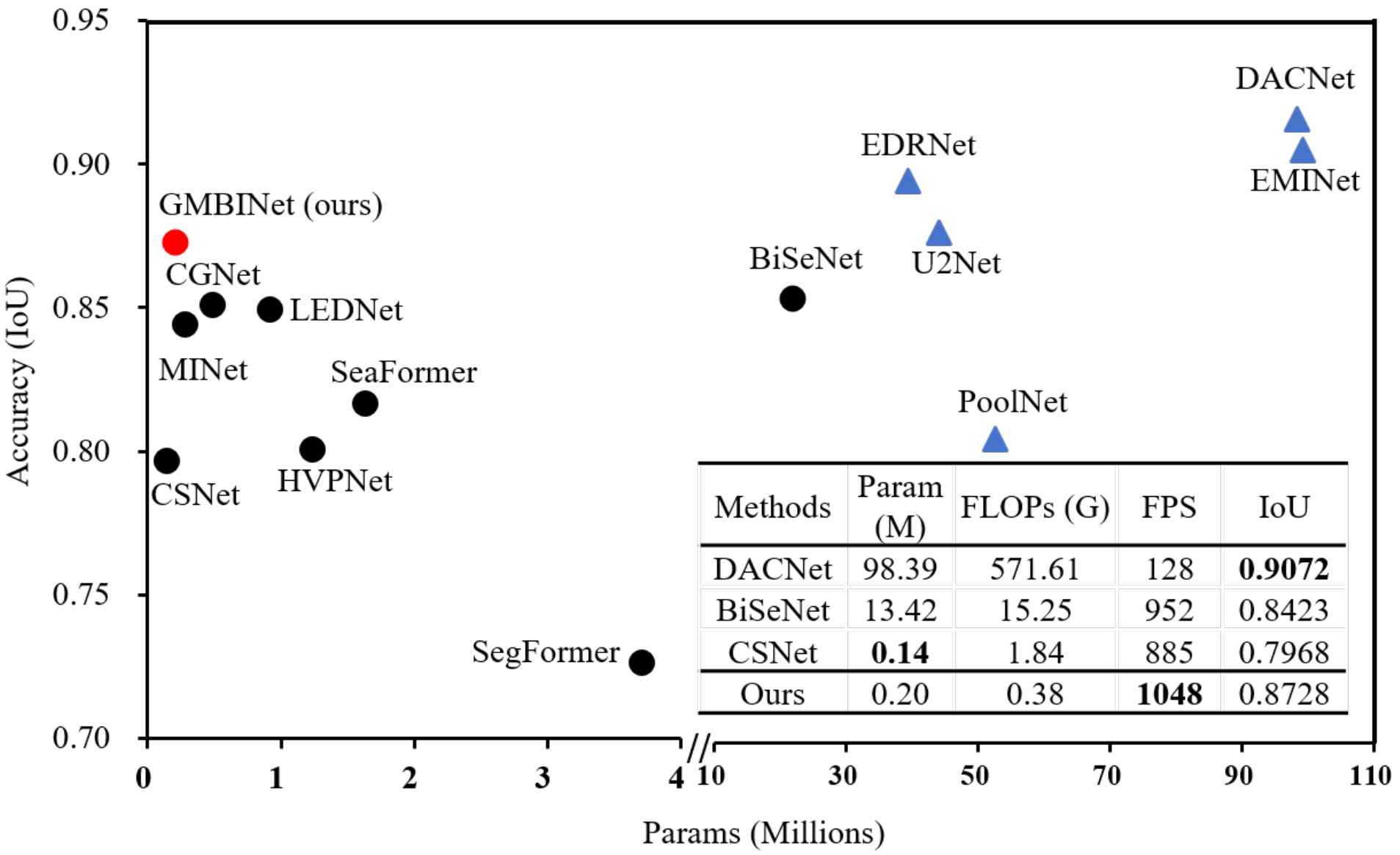}
    \caption{Comparison of defect detection accuracy and computational efficiency metrics between our proposed GMBINet and several state-of-the-art models on the SD-Saliency-900 dataset~\cite{song2020edrnet}. GMBINet achieves an optimal trade-off between detection performance and computational efficiency.}\label{fig: figure1}
\end{figure}

Building on the proposed GMBI module, we design a streamlined five-stage backbone and further develop GMBINet, a lightweight encoder–decoder network optimized for real-time detection of steel surface defects. GMBINet effectively balances accuracy and computational efficiency compared to other networks, as shown in Fig. \ref{fig: figure1}. The main contributions of this paper are as follows. 

\begin{enumerate}
    \item \textbf{GMBI module}: A lightweight multiscale module that integrates a group strategy and a bidirectional interaction mechanism for the efficient extraction and interaction of multiscale features.
    \item \textbf{BPFI mechanism and EWMS operation}: The BPFI mechanism integrates with the parameter-free EWMS operation to achieve progressive, cross-scale feature enhancement with negligible computational overhead.
    \item \textbf{GMBI-based backbone and GMBINet}: A streamlined backbone constructed by linearly stacking GMBI modules to reduce latency. Based on this, GMBINet is designed as a compact real-time encoder-decoder network optimized for steel surface defect detection, achieving a trade-off between accuracy and computational efficiency.
\end{enumerate}

\section{Related Work}
\label{sec:Related work}
\subsection{Surface Defect Detection}
Recent advances in DL have significantly enhanced surface defect detection due to superior feature extraction capabilities. These approaches are typically divided into two categories: bounding-box detection and pixel-wise detection. Bounding-box methods formulate defect detection as a localization problem, aiming to locate and classify defect regions by generating bounding boxes. For example, Huang et al. \cite{huang2025improved} proposed an improved YOLO model for the accurate detection of steel surface defects by incorporating a multilevel alignment module to alleviate misalignment of characteristics and a fusion redistribution module to integrate global contextual information. Chen et al. \cite{chen2024srpcnet} proposed the self-reinforcing perception coordination network to achieve more effective defect localization in seamless steel pipes by utilizing feature augmentation, hierarchical attention integration, and bilateral self-fusion. Cao et al. \cite{cao2024cacs} integrated a coordinate attention mechanism and a channel shuffle operation into the YOLOv8m framework to improve the detection accuracy of insulator defects. Although these methods demonstrate strong localization performance, they often fail to capture fine-grained morphological details, such as defect boundaries and topological structures, thereby limiting their applicability in closed-loop industrial analysis and quality control. To address these limitations, pixel-wise detection methods have been investigated, offering pixel-level localization that enables the extraction of detailed boundary and structural characteristics of defects. For instance, Song et al. \cite{song2020edrnet} introduced an encoder-decoder residual network (EDRNet) for accurate pixel-wise steel defect detection. Zhou et al. \cite{zhou2021dense} proposed a dense attention-guided cascaded network (DACNet), which combines multi-resolution branches with dense attention to ensure robust defect detection. Ma et al. \cite{ma2023sia} proposed a structural-aware model for pixel-wise defect detection that embeds realistic defects using a generative adjunctive network and aggregates spatial structure information via a self-attention encoder and spatial awareness decoder modules. Considering the advantages of pixel-wise approaches in accurately characterizing both the spatial location and morphological structure of defects, this study adopts a pixel-wise detection strategy for high-precision, real-time steel surface defect detection.

\subsection{Lightweight Real-Time Neural Network}
In recent years, lightweight real-time deep learning models have garnered considerable attention, facilitating the deployment of DL models in latency-sensitive and resource-constrained industrial environments \cite{gong2025real, han2020ghostnet, wu2020cgnet, gao2020highly, hou2025lightweight}. Within CNN-based frameworks, numerous lightweight architectures have been proposed to achieve a favorable balance between computational cost and representational capacity. Among them, MobileNet \cite{howard2017mobilenets} pioneered the use of DSConv to significantly reduce parameter size and computational complexity, laying the foundation for many subsequent lightweight and real-time models. Building upon this foundation, architectures such as ShuffleNet \cite{ma2018shufflenet} and CGNet \cite{wu2020cgnet} further enhance efficiency through channel splitting, group convolutions, and context-guided designs. BiSeNetV2 \cite{yu2021bisenet} integrates DSConv operations into its feature aggregation layers to facilitate feature fusion between the detail and semantic branches.  While CNN-based methods excel in leveraging local spatial priors and architectural simplicity, recent advances have introduced lightweight Transformer-based models that focus on capturing global dependencies with tolerable latency \cite{zhang2025tsd, ye2023real, xie2021segformer}. For example, MobileViT \cite{mehtamobilevit} combines local convolutions with compact Transformer blocks to enhance performance on mobile devices. EfficientFormer \cite{li2022efficientformer} replaces global attention with spatial-channel interactions, and FastViT \cite{vasu2023fastvit} employs token mixing along with convolutional embeddings to accelerate inference. SegFormer \cite{xie2021segformer} employs a multi-stage Transformer encoder without positional encodings and uses a lightweight decoder for efficient multiscale representation. SeaFormer \cite{wan2023seaformer} further improves efficiency through token-scale decoupled attention and scale-aware global context modeling. However, despite their strengths in global modeling, these Transformer-based models often require large-scale pretraining and lack strong spatial inductive biases, limiting their adaptability in data-scarce or latency-critical scenarios \cite{han2022survey}. Considering these trade-offs, we adopt a CNN-based framework as the foundation of our real-time segmentation model. Specifically, we extend the lightweight design of DSConv and incorporate it as the core component of our proposed GMBI module, which is designed to meet the demands of real-time dense prediction with improved multiscale representation capabilities.

\subsection{Multiscale Learning}
Objects in industrial scenes often exhibit significant variations in scale, posing considerable challenges for dense prediction tasks that rely solely on single-scale representations. To address this, various multiscale feature extraction strategies have been developed. In CNN-based frameworks, methods such as DeepLabV3+ \cite{chen2018encoder} and TridentNet \cite{li2019scale} utilize parallel dilated convolutions with varying dilation rates to capture contextual information from multiple receptive fields. Other approaches, including DDRNet \cite{pan2022deep}, PoolNet++ \cite{liu2022poolnet+}, and P2T \cite{wu2022p2t}, employ multi-stride pooling to efficiently gather features across spatial scales. More recently, Transformer-based models have extended multiscale modeling into the token space. SegFormer \cite{xie2021segformer} incorporates a multi-stage encoder with scale-adaptive representations and discards positional encodings to enhance generalization. PVT \cite{wang2021pyramid} employs a hierarchical pyramid structure, while HiViT \cite{yu2025hivit} integrates hierarchical pooling to generate cross-resolution semantic features. Despite their success, most of these models entail high computational costs and lack effective multiscale interaction. Although various attention-based approaches have been proposed to enhance cross-scale interaction, they typically introduce additional computational overhead \cite{zeng2024multiscale, song2025dusa}. To mitigate both over-parameterization and scale isolation, MINet \cite{shen2024minet} proposes a multiscale strategy using DSConv with varying dilation rates. It reorganizes multiscale features along the channel dimension and applies pointwise convolutions for efficient channel-wise interaction. However, its independent multibranch design and post-hoc fusion mechanism lead to computational redundancy and limit dynamic interaction between scales during feature encoding. To overcome these limitations, we propose the GMBI module, which performs group-wise multiscale extraction and facilitates interaction across scales in a bidirectional progressive manner. This design facilitates fine-grained multiscale representation without increasing computational complexity, making it particularly suitable for resource-constrained visual applications.

\section{Methodology}
\label{sec:methods}
\subsection{Preliminary and Motivation}
\begin{figure}
  \centering
    \includegraphics[width=0.9\linewidth]{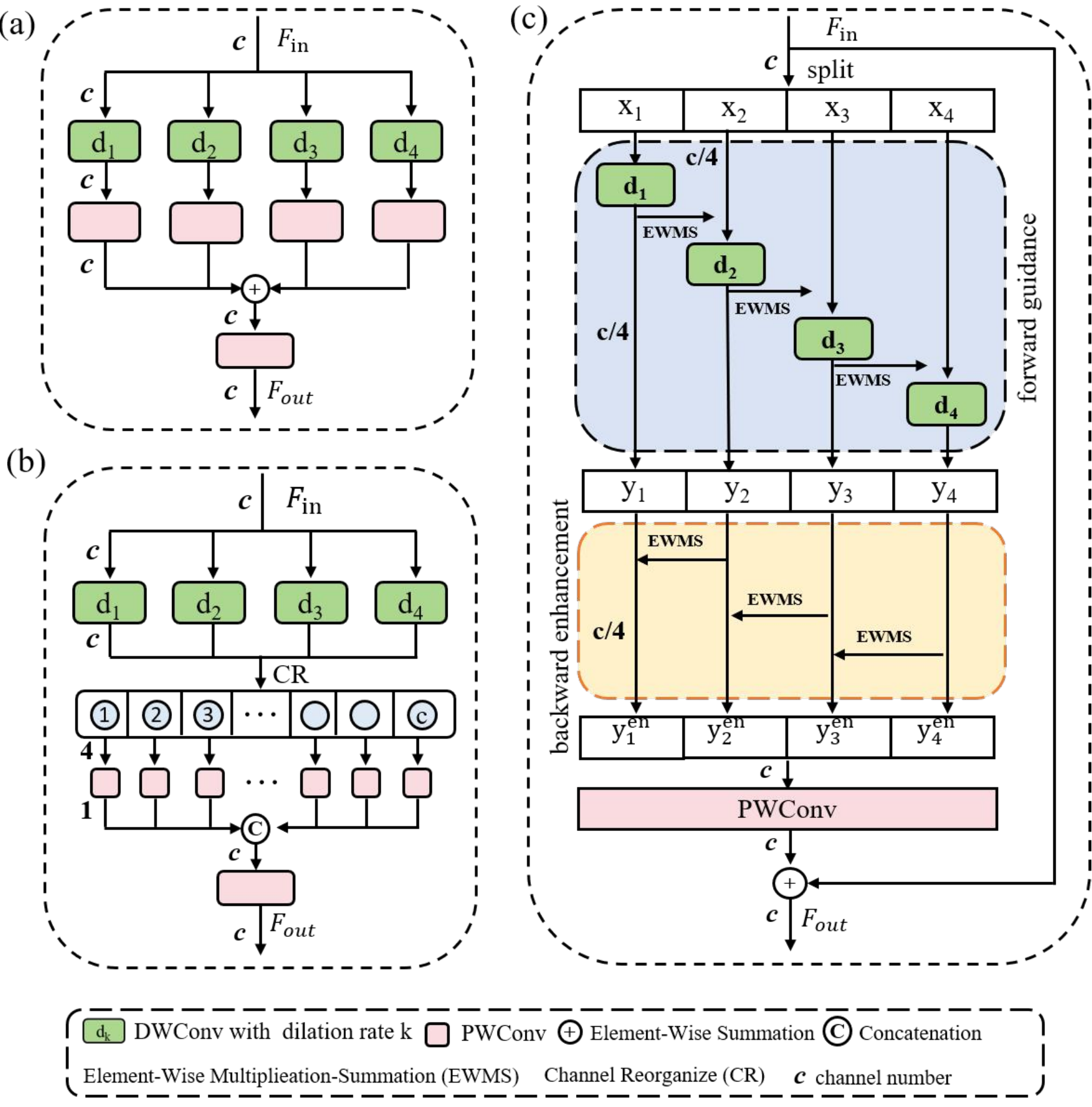}
    \caption{Comparison of different lightweight multiscale feature extraction modules (the scale dimension $n = 4$). (a) DSConv-based lightweight multibranch module, where multiscale feature maps are independently extracted in a multibranch manner without cross-scale interaction. (b) MI Module, where multiscale feature maps are independently extracted in a multibranch manner, followed by post-hoc channel-wise interactions across scales via c PWConv operations. (c) GMBI Module, where multiscale features are extracted in a group-wise manner and interact through a parameter-free, cost-efficient bidirectional progressive mechanism. The numbers or the letter c labeled beside the lines indicate the number of channels in the input or output feature maps.}\label{fig: MultiBranch}
\end{figure}

In this subsection, we review the composition and computational complexity of the DSConv. As illustrated in Fig.~\ref{fig: MultiBranch}(a), given an input tensor $F_{in} \in \mathbb{R}^{c \times w \times h}$ with channel number $c$, width $w$, and height $h$, a typical DSConv can replace a standard convolution in two steps. First, a $k\times k$ depthwise convolution (DWConv) is utilized to extract spatial information within each channel. Then, a pointwise convolution (PWConv) operation is used to model channel interconnections. The total computational cost of DSConv can be expressed as:
\begin{equation}
k^2\times c \times h \times w+c^2 \times h\times w.\label{eq-dsconv}
\end{equation}

Moreover, to achieve lightweight multiscale feature extraction, various methods have been proposed, with a prevalent trend to adopt DSConv-based multibranch architectures. As shown in Fig.~\ref{fig: MultiBranch}(a), the input features are processed through a series of DSConv blocks with varying configurations to extract multiscale features, which are then fused by a summation operation and aggregated using a cost-efficient $1 \times 1$ convolution block. For convenience, we define $n$ as a control parameter for the scale dimension, where a larger $n$ represents the inclusion of more multiscale levels. Therefore, the computational cost of the above multiscale feature extraction process can be calculated as follows:
\begin{equation}
n \times (k^2 \times c \times h \times w + c^2 \times h \times w) + c^2 \times h\times w.\label{multi-branch}
\end{equation}
While effective,  this approach incurs a higher computational cost and limits inter-scale information exchange. To address these, Shen et al. \cite{shen2024minet} proposed a lightweight multiscale interactive (MI) module, which first employs DWConv operations in a multibranch manner for initial multiscale feature extraction. Subsequently, the multiscale features are reorganized along the channel dimension to produce $c$ feature maps of size $n \times h \times w$ by aggregating the corresponding channels on different scales. This is followed by channel-wise interactions across scales, implemented via $c$ PWConv operations. Finally, an additional PWConv is used to integrate features along the channel dimension. The computational cost of the MI module, illustrated in Fig.~\ref{fig: MultiBranch}(b), is calculated as:
\begin{equation}
n \times (k^2 \times c \times h \times w) + c \times(n \times h \times w) + c^2 \times h \times w. \label{EqMINet}
\end{equation}
Despite its improvements in scale-aware learning, the MI module still inherits the linear complexity growth of multibranch designs (see Eq. \ref{EqMINet}) and lacks sufficient scale interaction. In particular, its post-hoc feature interaction fails to effectively guide the multiscale feature extraction process, limiting its multiscale representation capability.

These limitations underscore the need for a more efficient and integrated solution that can simultaneously achieve multiscale feature extraction and cross-scale interaction, while maintaining low computational cost. To this end, we propose the GMBI module, which combines a group multiscale extraction strategy with a lightweight bidirectional interaction mechanism. The architecture of the GMBI module is detailed in the following subsection.

\subsection{GMBI Module}
\subsubsection{Architecture} As depicted in Fig.~\ref{fig: MultiBranch}(c), GMBI contains four components: group multiscale feature extraction, bidirectional progressive feature interactor, element-wise multiplication-summation operation, and multiscale feature fusion. Among them, the first three form the core of GMBI, collectively designed to enable efficient and lightweight multiscale feature representation and interaction.

\textbf{Group Multiscale Feature Extraction}. 
To reduce the redundancy of feature maps and facilitate the design of a lightweight model, our GMBI module extracts multiscale features in a group manner, as shown in Fig.~\ref{fig: MultiBranch}(c). Specifically, the GMBI module evenly splits the input tensor ${F_{in}} \in \mathbb{R}^{c \times w \times h}$ into $n$ equal-channel feature subsets, denoted by ${x_i} \in \mathbb{R}^{\frac{c}{n} \times w \times h}$, $i \in \{1, 2, ..., n\}$. We employ DWConv with varying dilation rates (i.e., 1, 2, ..., $n$) to extract multiscale features from different subsets. Notably, this group-wise design retains the efficiency of DWConv while enabling multiscale feature modeling. Unlike multibranch structures that incur rising complexity, it improves representational capacity without additional overhead.

\textbf{Bidirectional Progressive Feature Interactor (BPFI)}. 

The design of the Bidirectional Progressive Feature Interactor (BPFI) is inspired by neurocognitive theories of visual perception, where visual understanding emerges from the dynamic interplay between bottom-up sensory input (from lower-level details to higher-level semantics) and top-down cognitive modulation (from larger-scale context to smaller-scale enhancement) \cite{hochstein2002view, friston2010free, kersten2004object}. As illustrated in Fig.3(c), BPFI explicitly instantiates this paradigm via two complementary components: forward guidance and backward enhancement. Specifically, in the forward guidance process, smaller-scale feature maps act as guidance cues to assist neighboring scale groups in extracting higher-level features. This simulates the bottom-up integration of details into higher-level semantics. Conversely, in the backward enhancement process, higher-level and larger-scale features provide modulation signals to refine adjacent lower-level and smaller-scale features, mirroring top-down contextual modulation.

To formalize the BPFI mechanism, given a set of input subsets $\{x_1, x_2, \ldots, x_n\}$, the forward guidance generates intermediate multiscale features $\{y_1, y_2, \ldots, y_n\}$ as:
\begin{equation}
y_i=
\begin{cases}
f^{i}_{DS}(x_i)& \text{$i = 1$}\\

f^{i}_{DS}(f_{inter}(y_{i-1}, x_i))& $ i > 1$\\
\end{cases}.
\end{equation}
Here, $f^i_{DS}$ and $f_{inter}$ denote a $3 \times 3 $ DSConv with dilation rate $i$ and multiscal interaction operation, respectively. And then, the backward enhancement stage further refines these features to obtain the final multiscale outputs $\{y^{en}_{1}, y^{en}_{2}, \ldots, y^{en}_{n}\}$ via:
\begin{equation}
y_i^{en}=
\begin{cases}
f_{inter}(y_{i}, y_{i-1}) & \text{$ i > 1$}\\
y_i& \text{$i = n$}\\
\end{cases}.
\end{equation}
These refined multiscale features are subsequently aggregated within the GMBI module to generate the final output.

\begin{figure}
    \centering
    \includegraphics[width=0.70\linewidth]{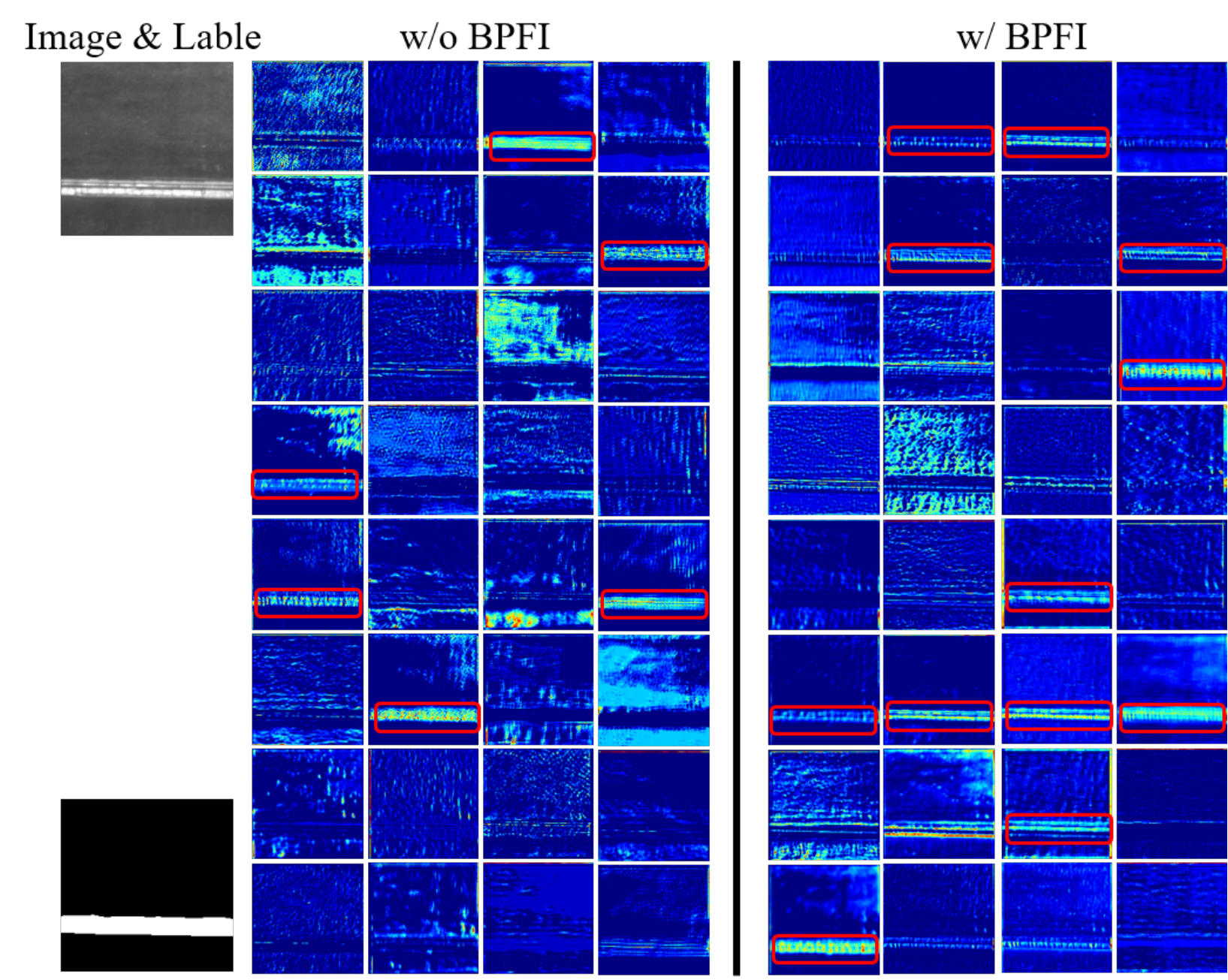}
    \caption{Visualization of feature maps generated by the last GMBI module in the second stage of GMBINet. Each column represents features extracted at different scales, while each row corresponds to different channels within the corresponding groups. Warmer colors indicate higher attention coefficient values. Feature maps on the left are from models without BPFI, while those on the right are from models with BPFI. Salient target regions are highlighted with red bounding boxes.
}
    \label{fig:BPFI_heatmap}
\end{figure}

Overall, BPFI provides four key advantages. (1) \textbf{Robust multiscale modeling}: Bidirectional interaction enables contextual enhancement across scales (see Fig. \ref{fig:BPFI_heatmap} (c)), enhancing feature discrimination without introducing additional parameters. (2) \textbf{Efficient receptive field expansion}: The progressive accumulation allows each DWConv layer to inherit contextual information from preceding subsets \cite{gao2019res2net}, thereby expanding the effective receptive field and enhancing high-level semantic encoding. (3)\textbf{ Improved semantic consistency}: Bidirectional refinement enforces feature coherence, facilitating stable convergence and efficient training \cite{wang2020deep}, which is beneficial for efficient deployment. (4) \textbf{Modular flexibility}: The cognition-inspired structure enables BPFI to be seamlessly integrated into various modules, supporting scalable deployment in practical applications.

\begin{figure}
    \centering
    \includegraphics[width=0.50\linewidth]{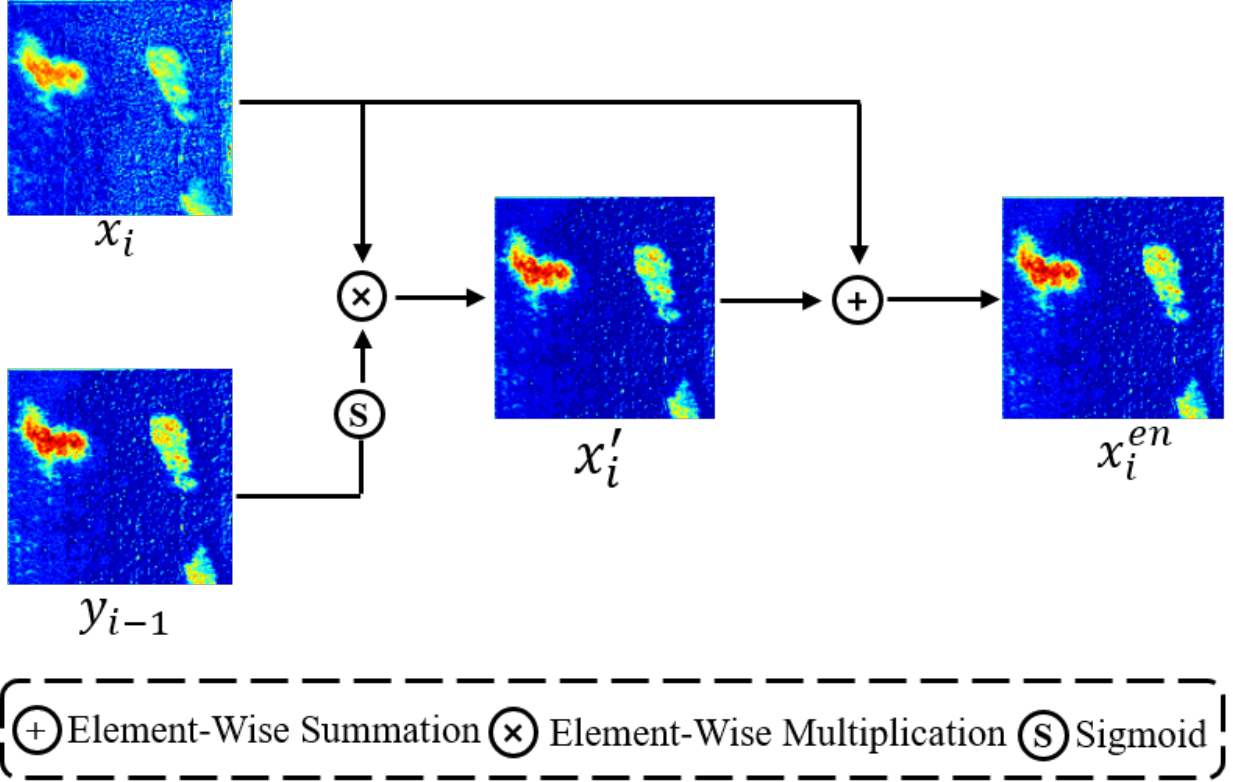}
    \caption{Visualization of EWMS operation. Warmer colors represent higher attention coefficient values. First, an element-wise multiplication operation between $y_{i-1}$ and $x_i$ is applied to enhance the perception of the target regions in $x_i$, resulting in $x^{'}_i$. Then, an element-wise summation operation with $x_i$ is performed to mitigate information loss, resulting in $x^{en}_i$.}
    \label{fig:EWMS}
\end{figure}

\textbf{Element-Wise Multiplication-Summation}. To facilitate a cost-efficient multiscale feature interaction, we propose a novel EWMS operation as the implementation of the interaction function
$f_{inter}$ within the BPFI mechanism. Although element-wise summation is widely used for its simplicity, it often leads to semantic dilution and limited noise suppression \cite{xia2024cross}. In contrast, element-wise multiplication can emphasize target regions by suppressing irrelevant activations,  but this can come at the cost of losing fine-grained details \cite{ma2024rewrite}. Although feature concatenation can retain both semantic and detailed information, it typically requires additional convolutional layers for feature fusion and channel alignment, leading to an increased computational burden. To address these trade-offs, EWMS integrates the advantages of both multiplication and summation within a parameter-free operation. As shown in Fig. \ref{fig:EWMS}, in the forward guidance process, given the guidance feature maps $y_{i-1}$ and the feature maps to be processed $x_i$, an element-wise multiplication operation is first applied to improve the perception of the target regions in $x_i$, resulting in $x^{'}_i$. This is followed by an element-wise summation operation with $x_i$ to mitigate information loss,  resulting in $x^{en}_i$. Thus, the overall process of the EWMS operation can be summarized as follows:
\begin{equation}
x_i^{en}= \sigma(y_{i-1}) \times  x_{i} +  x_{i},
\label{EWMS}
\end{equation}
where $\sigma$ denotes the sigmoid function. A similar process is applied in the backward path, where higher-level features guide the refinement of lower-level representations. Crucially, EWMS introduces no additional parameters or convolutional operations, making it a lightweight yet effective operation well-suited for real-time applications where both accuracy and efficiency are critical.

\textbf{Multiscale Feature Fusion}. The enhanced feature maps $\{y^{en}_{1}, y^{en}_{2}, \ldots, y^{en}_{n}\}$ are then concatenated for channel restoration and refined using a PWConv to establish interconnections along the channel dimension. Moreover, we incorporate a residual connection to combine the original input tensor $F_{in}$, resulting in the final output of our GMBI module, denoted as $F_{out}$. The complete process can be written as:
\begin{equation}
F_{out} = F_{in} + f_{pw}(cat(y^{en}_{1}, y^{en}_{2}, ... , y^{en}_{n})),\label{eq}
\end{equation}
where $f_{pw}$ denotes PWConv, and $cat$ represents concatenate operation.

\subsubsection{Analysis of Complexities} According to Fig. \ref{fig: MultiBranch}(c)  and the descriptions above, the computational cost of our GMBI module can be calculated as follows: 
\begin{equation}
n \times ({k}^2 \times \frac{c}{n} \times h\times w) + c^2 \times h\times w.\label{GMSI-a}
\end{equation}
Compared to Eq. \ref{GMSI-a} and Eq. \ref{eq-dsconv}, the computational complexity of the proposed GMBI module is equivalent to that of DSConv. To further demonstrate its efficiency, we compare GMBI with conventional multibranch approaches. From Eq. \ref{multi-branch}, it is evident that multibranch methods suffer from significant complexity accumulation effects. As more scales are considered, the overall complexity of the model increases linearly as more feature channels need to be processed. In contrast, the GMBI module achieves scale-invariant complexity because the number of channels in the feature maps remains fixed, regardless of the number of scales considered, as illustrated in Fig. \ref{fig: GMBINet}. Moreover, compared to the recently proposed MI module \cite{shen2024minet}, the GMBI achieves efficient multiscale feature interaction in a bidirectional manner by incorporating forward guidance during extraction and backward enhancement after extraction, without introducing additional computational overhead. Based on the above analysis, we can confirm that the GMBI module establishes a superior lightweight architecture for multiscale feature extraction and interaction compared to existing approaches.

\subsection{Overall Architecture}

Most existing surface defect detection methods rely on classical backbones such as ResNet for semantic feature extraction, but suffer from high computational and storage overhead. To address these limitations, we propose a lightweight backbone built upon the GMBI module, which is then integrated into a conventional encoder-decoder framework for efficient pixel-wise defect detection.

\subsubsection{GMBI-Based Lightweight Backbone}
\begin{figure}
    \centering
    \includegraphics[width=1.0\linewidth]{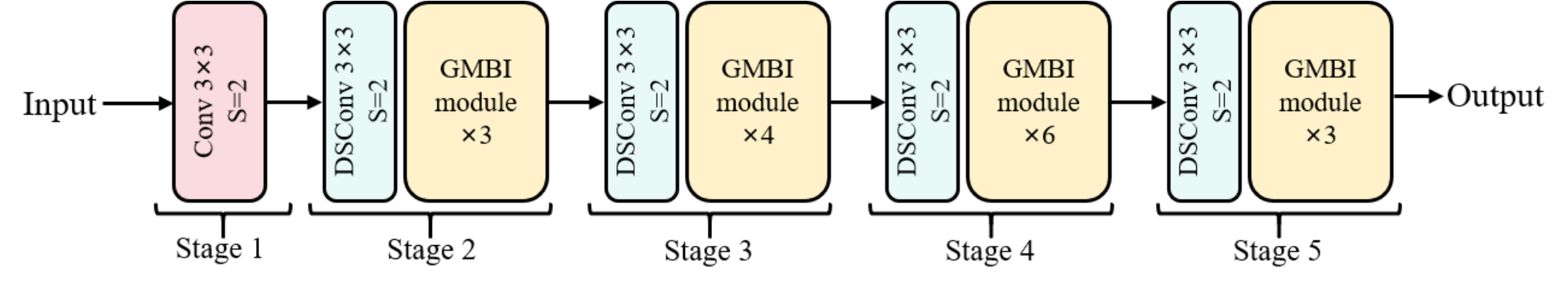}
    \caption{Illustration of the GMBI-based backbone. A five-stage backbone with GMBI modules sequentially arranged in the last four stages.}
    \label{fig: GMBIBackbone}
\end{figure}

GMBI-Based backbone follows a five-stage architecture similar to ResNet, as shown in Fig. \ref{fig: GMBIBackbone}. The first stage uses a $3 \times 3$ convolution with a stride of 2 for initial feature extraction. The remaining four stages comprise stacked GMBI modules that progressively extract high-level semantics with reduced latency. Between stages, a DSConv with a stride of 2 performs spatial downsampling and channel expansion. Inspired by ResNet-50, stages 2 to 5 contain 3, 4, 6, and 3 GMBI units, respectively. The details of our GMBI-based backbone are presented in Table \ref{tab:backbone}.

\begin{table}[!t]
\begin{center}
\caption{Details of the GMBI-based backbone configuration, where \#M and \#F represent the number of modules and channels, respectively. }
\label{tab:backbone}
\begin{tabular*}{\tblwidth}{@{}CCCCCC@{}}
\toprule
\textbf{Stage} & \textbf{Resolution} & \textbf{Module} & \textbf{\#M} & \textbf{\#F} \\
\midrule
\multirow{2}{*}{1}&512×512&\multirow{2}{*}{Conv $3\times 3$}&\multirow{2}{*}{1}&\multirow{2}{*}{16}\\
&256×256& & &\\
\bottomrule
\multirow{2}{*}{2}&256×256& DSConv $3\times 3$&1&32\\
&128×128&GMBI module& 3&32\\
\bottomrule
\multirow{2}{*}{3}&128×128& DSConv $3\times 3$&1&64\\
&64×64&GMBI module& 4&32\\
\bottomrule
\multirow{2}{*}{4}&64×64& DSConv $3\times 3$&1&96\\
&32×32&GMBI module& 6&96\\
\bottomrule
\multirow{2}{*}{5}&32×32& DSConv $3\times 3$&1&128\\
&16×16& GMBI module& 3&128\\
\bottomrule
\end{tabular*}
\end{center}
\end{table}

\subsubsection{GMBINet Architecture}
\begin{figure}
    \centering
    \includegraphics[width=0.7\linewidth]{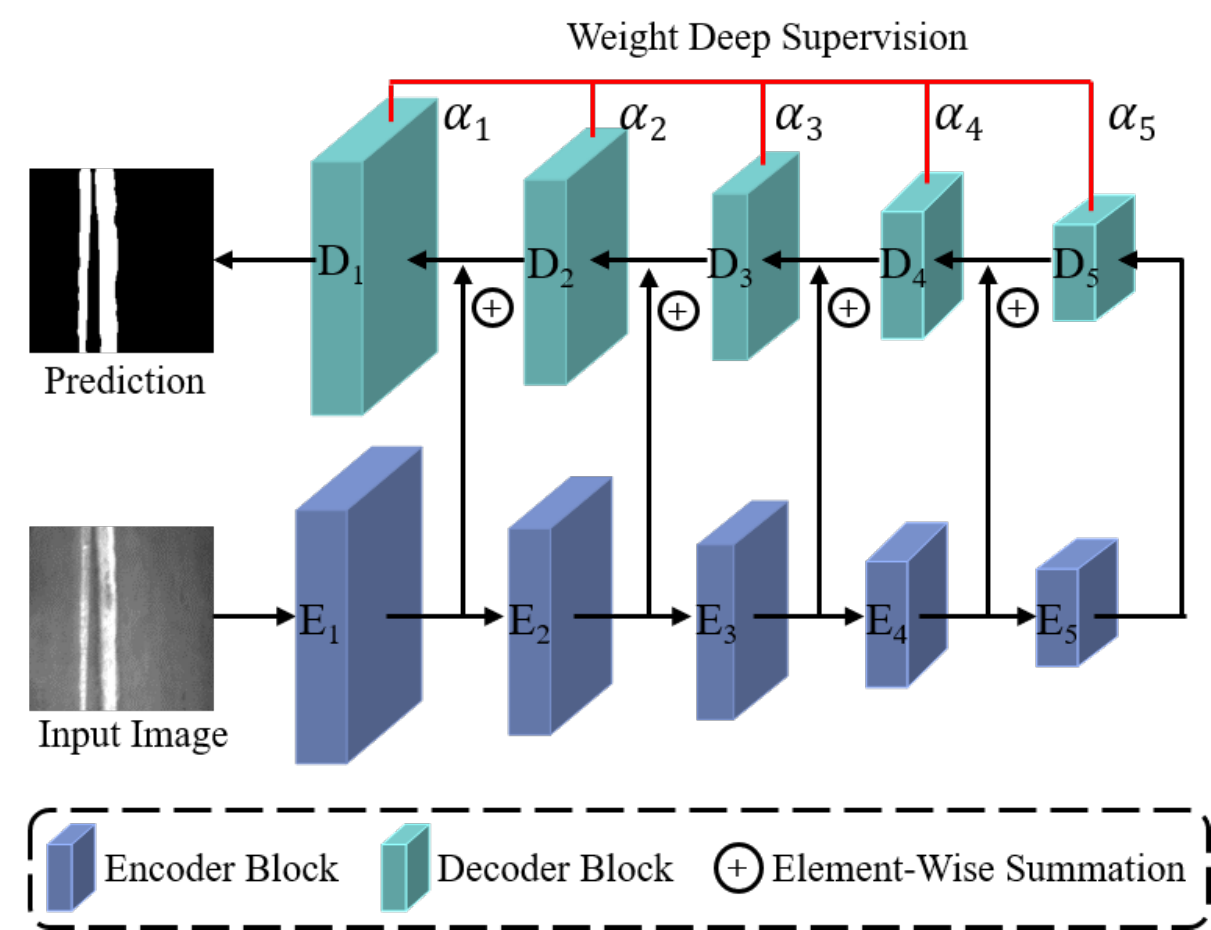}
    \caption{Illustration of proposed GMBINet. GMBINet is a 5-stage encoder-decoder architecture featuring the GMBI-based real-time backbone for multilevel feature extraction used in the encoder.}
    \label{fig: GMBINet}
\end{figure}

Based on the GMBI-based backbone, we proposed a lightweight detection model (named GMBINet) for the real-time detection of strip steel surface defects. As shown in Fig.~\ref {fig: GMBINet}, GMBINet is implemented in the classical 5-stage encoder-decoder architecture. In the encoder, the aforementioned GMBI-based backbone is utilized to extract multilevel semantic features.  Each of these feature maps has a resolution of $E_i \in \mathbb{R}^{C_i \times \frac{H}{2^{i}} \times \frac{W}{2^{i}}}$, where $i=\{1,2,3,4,5\}$. Here, $C_i$ denotes the number of channels in the $i$-th stage of the encoder, $H$ and $W$ refer to the height and width of the input images, respectively. In the decoder, progressive up-sampling is employed to gradually recover feature maps to the input image size. Each decoder includes a bilinear interpolation operation for feature map up-sampling and a DSConv operation for channel dimension adjustment, resulting in a decoder feature map with a resolution of $D_i \in \mathbb{R}^{C_i \times \frac{H}{2^{i}} \times \frac{W}{2^{i}}}$, where $i=\{1,2,3, 4,5\}$. 

\subsection{Loss Function}
 We combine the binary cross entropy (BCE) loss and the structure similarity index measure (SSIM) loss as the basic loss for the network. For the annotation label $y$ and the predictions $\hat{y}$, the proposed hybrid loss function can be expressed as follows:
 \begin{equation}
 L=L_{BCE}(y,\hat{y})+L_{SSIM}(y,\hat{y}).
 \end{equation}

where $L_{BCE}$ and $L_{SSIM}$ denote the BCE loss and SSIM loss respectively. In addition, a deep weighted supervision strategy is employed for model optimization. The final loss is calculated as the weighted sum of the individual losses from each encoder stage, summarized as follows:
 \begin{equation}
L_{total}= \alpha_1 L_1+\alpha_2 L_2+\alpha_3 L_3+\alpha_4 L_4+\alpha_5 L_5,
\end{equation}
where $L_1$ to $L_5$ denote the losses at each stage of the decoder, and $\alpha_1$ to $\alpha_5$ are the weighted coefficients assigned to the corresponding losses, and both are set to 1.0 following \cite{shen2024minet}.

\section{Experiments}
\label{sec:experiments}
\subsection{Dataset and Evaluation Metrics}
We evaluated the performance of the proposed method on two challenging surface defect detection datasets: the SD-Saliency-900~\cite{song2020edrnet} dataset for strip steel surface defects and the NRSD-MN \cite{zhang2020mcnet} dataset for rail surface inspection. \textbf{SD-Saliency-900} consists of 900 grayscale images (200 × 200 resolution) with three types of defects (i.e., inclusion, patches, and scratches), and each type comprises 300 images with pixel-wise annotations. Following the dataset splits in \cite{song2020edrnet, zhou2021dense}, the standard training set comprises 810 images, with 540 randomly selected defect images (180 per type) from SD-Saliency-900 and 270 images (90 per type) disturbed by salt-and-pepper noise. \textbf{NRSD-MN} dataset comprises 3936 RGB images with varying resolutions. Following the dataset division in \cite{zhang2020mcnet}, 2086 images were used for model optimization, 885 for validation, and 965 for testing.

To quantitatively evaluate the performance of the proposed method, we utilize five commonly employed metrics in strip steel surface defect detection: mean absolute error (MAE), weighted F-measure (WF), overlapping ratio (OR), structure-measure (SM), and Pratt’s figure of merit (PFOM). Additionally, computational efficiency is evaluated using three indicators: parameter size (Param), floating-point operations (FLOPs), and inference speed measured in frames per second (FPS).

Furthermore, we evaluated the classification performance of the proposed method on the \textbf{NEU-CLS} \cite{song2013noise} dataset, which consists of 1800 grayscale images with six types of defects (i.e., rolled-in scale, patches, crazing, pitted surface, inclusion, and scratches), and each category includes 300 images. To ensure a class-balanced evaluation, 70$\%$ of the images from each class were randomly selected for training, 10$\%$ for validation, and the remaining 20$\%$ for testing. For classification performance, we report four standard metrics: accuracy, precision, recall, and F1-score.

\subsection{Implementation Details}
We conducted all experiments on hardware with an NVIDIA GeForce GTX 4090 GPU (24 GB) and a Gen Intel(R) Core (TM) i9 13900KF 3.00 GHz. All networks were trained for 50,000 iterations with the Adam optimizer until convergence. The initial learning was set to $4\times {10}^{-3}$ and adjusted according to a cosine annealing schedule. The batch size was configured to 32, and all images were uniformly resized to 512×512 pixels in the training phase. 
In addition, to avoid overfitting, data augmentation techniques such as z-score normalization, random flipping, intensity shifts, and scaling were applied during training to improve sample diversity. During the inference phase, images were also resized to 512 × 512, and saliency results were further restored to their original resolution using bilinear interpolation.

\subsection{Comparison Experiment}
\subsubsection{Strip Steel Surface Defects Detection}
In this section, we compare our GMBINet with 14 state-of-the-art models, including six traditional DL-based approches (i.e., PoolNet \cite{liu2019simple}, U2Net \cite{qin2020u2}, EDN \cite{wu2022edn},  EDRNet \cite{song2020edrnet}, DACNet \cite{zhou2021dense}, EMINet \cite{zhou2021edge}), and eight lightweight real-time methods (i.e., BiSeNet \cite{yu2018bisenet}, LEDNet \cite{wang2019lednet}, CGNet \cite{wu2020cgnet}, CSNet \cite{gao2020highly}, MINet \cite{shen2024minet}, HVPNet \cite{liu2020lightweight},  SegFormer \cite{xie2021segformer}, SeaFormer \cite{wan2023seaformer} ).

\textbf{Quantitative Results}. As shown in Table \ref{tab:SOTA}, the proposed GMBINet achieves competitive performance with the highest GPU inference speed (1048 FPS), and the lowest computational complexity (Param: 0.19 M, FLOPs: 0.39 G) compared to traditional methods. For instance, the SM value of the top-performing DACNet is 0.9417, while our GMBINet’s SM value is 0.9334 ($\downarrow$ 0.88$\%$). However, the parameters and FLOPs of our GMBINet are only 0.19$\%$ (0.19 M vs 98.39 M) and 0.07$\%$ (0.39 G vs 571.61 G) of DACNet, respectively, while also being $7.19 \times$ faster in GPU inference speed (1048 FPS vs 128 FPS). Considering the potential scarcity of GPU resources in real-world industrial scenarios, we further evaluate the inference speeds of both GMBINet and DACNet under CPU-only conditions. GMBINet achieves 16.53 FPS, significantly outperforming DACNet, which reaches only 0.89 FPS under the same settings. These results highlight the substantial efficiency advantage of GMBINet over DACNet and underscore its strong potential for real-time deployment on resource-constrained edge devices. Furthermore, compared to lightweight real-time approaches, our GMBINet outperforms all compared methods across all quantitative evaluation metrics (i.e., MAE, WF, OR, SM, PFOM, and IoU), while achieving the fastest speed, lowest FLOPs, and the second lowest parameter size. For example, while BiSeNet achieved the fastest speed (952 FPS) among the three representative real-time methods, GMBINet further achieves a 10.08 $\%$ improvement in inference speed. Moreover, compared to MINet, which achieves the best performance among the three lightweight methods, our GMBINet shows superior detection performance (e.g., MAE: 0.0169 vs 0.0126, OR: 0.8115 vs 0.8653), lower computational complexity (e.g., Param: 0.19 M vs 0.28 M, FLOPs: 0.39 G vs 0.64 G), and faster inference speed (1048 FPS vs 690 FPS). Although CSNet slightly outperforms our method in terms of parameter size (0.14 M vs 0.19 M), its detection performance (e.g., SM: 0.8839 vs 0.9934, IoU: 0.8207 vs 0.8977) and inference speed (885 FPS vs 1048 FPS) are considerably lower than those of our GMBINet. 

\begin{table}[!t]
\begin{center}
\caption{Quantitative comparison of our GMBINet with 12 state-of-the-art methods on the SD-Saliency-900 dataset.The best three results in each column have been highlighted in red, green, and blue, respectively.}
\label{tab:SOTA}
\begin{tabular*}{\tblwidth}{@{}CCCCCCCCCCCC@{}}
\toprule
 Methods
    & \makecell{Param\\(M)↓}
    & \makecell{FLOPs\\(G)↓}
    & \makecell{Speed\\(FPS)↑}
    & MAE↓
    & WF↑
    & OR↑
    & SM↑
    & PFOM↑
    & IoU↑\\
\midrule
PoolNet\cite{liu2019simple}&52.51&254.79&139&0.0215&0.8476&0.7456&0.9022&0.8649&0.8058\\
U2Net \cite{qin2020u2} &44.01&150.61&167&0.0143&0.9071&	0.8134&0.9295&0.9088&0.8777\\
EDN \cite{wu2022edn} &21.83&148.73&123&0.0149&0.9115&	0.8308&0.9240&0.9021&0.8549\\
\hline
EDRNet \cite{song2020edrnet} &39.31&169.00&156	&0.0130& 0.9225& 0.8417&\textcolor{blue}{\textbf{0.9375}}&\textcolor{blue}{\textbf{0.9133}}&\textcolor{blue}{\textbf{0.8958}}\\ 
DACNet \cite{zhou2021dense}&98.39&571.61&128&\textcolor{red}{\textbf{0.0118}}&\textcolor{red}{\textbf{0.9275}}&	\textcolor{green}{\textbf{0.8464}}&\textcolor{green}{\textbf{0.9417}}&\textcolor{red}{\textbf{0.9204}}&
\textcolor{red}{\textbf{0.9072}}\\ 
EMINet \cite{zhou2021edge}&99.13&559.46&117&\textcolor{green}{\textbf{0.0119}}&\textcolor{blue}{\textbf{0.9253}}& 	\textcolor{blue}{\textbf{0.8447}}&\textcolor{red}{\textbf{0.9422}}&\textcolor{green}{\textbf{0.9173}}&
\textcolor{green}{\textbf{0.9064}}\\ 
\hline
BiSeNet \cite{yu2018bisenet}&13.42&15.25&\textcolor{green}{\textbf{952}}&0.0178&	0.8803&0.7848&0.9155&0.8868&0.8423\\
LEDNet \cite{wang2019lednet} &0.92&5.72&425&0.0182&0.8782&	0.7730&0.9196&0.8929&0.8495\\ 
CGNet \cite{wu2020cgnet}&0.49&3.55&639&0.0172&0.8825& 	0.7854&0.9191&0.8906&0.8509\\ 
\hline
CSNet \cite{gao2020highly} &\textcolor{red}{\textbf{0.14}}&\textcolor{blue}{\textbf{1.84}}&\textcolor{blue}{\textbf{885}}&0.0309&0.7659&0.7026&0.8839&0.8443&0.7968\\ 
HVPNet \cite{liu2020lightweight}&1.24&2.69&655&0.0271&0.8231&0.7110&0.8938&0.8442&0.8005\\ 

MINet \cite{shen2024minet} &\textcolor{blue}{\textbf{0.28}}&\textcolor{green}{\textbf{0.64}} 
&690&0.0169&0.8945&0.8115&0.9207&0.8943&0.8441\\ 

SegFormer \cite{xie2021segformer}& 3.71&6.76&404&0.0346&0.8241&0.7168&	0.8466&0.7517&0.7264 \\
SeaFormer \cite{wan2023seaformer}& 1.63&2.35&945&0.0188&0.8928&0.8075&	0.8997&0.8561&0.8164\\
\hline
Ours&\textcolor{green}{\textbf{0.19}}&\textcolor{red}{\textbf{0.39}}&\textcolor{red}{\textbf{1048}}&\textcolor{blue}{\textbf{0.0126}}&\textcolor{green}{\textbf{0.9273}}&\textcolor{red}{\textbf{0.8653}}&0.9334&0.9077&0.8728\\ 
\bottomrule
\end{tabular*}
\end{center}
\end{table}

\begin{figure}   
    \centering    
    \includegraphics[width=0.95\linewidth]{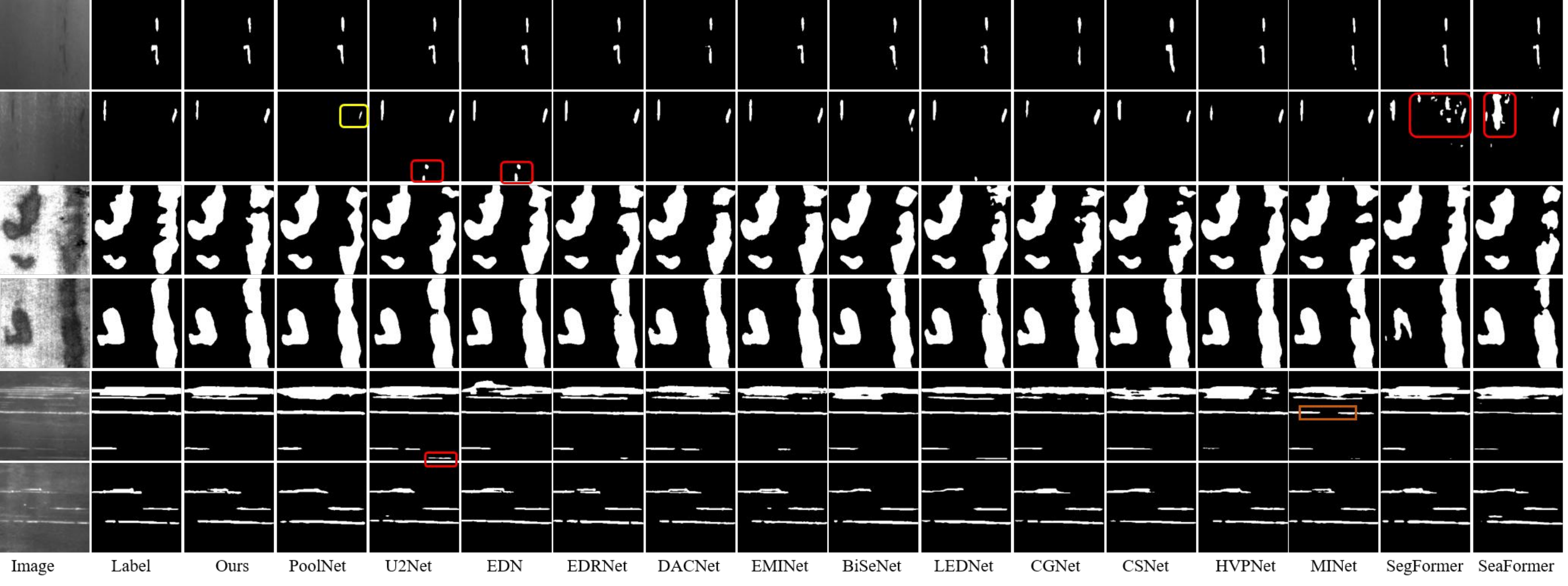}
    \caption{Qualitative results for different methods on SD-Saliency-900 dataset. The yellow box highlights under-segmentation areas in PoolNet, while the red boxes denote over-segmentation areas in U2Net, EDN, SegFormer, and SeaFormer. The orange box marks disconnected regions in MINet.}
    \label{fig:STOA}
\end{figure}

\textbf{Qualitative Results}. As shown in Fig. \ref{fig:STOA}, GMBINet achieves comparable performance to approaches designed for common natural scene images and surface defect detection. Specifically, GMBINet delivers significantly more accurate results in defect regions, whereas PoolNet, U2Net, and EDRNet suffer from severe over-segmentation (highlighted in the red box) and under-segmentation (highlighted in the yellow box). This superior performance is largely attributed to GMBINet's efficient multiscale feature extraction and interaction mechanism, which enhances the model's scale awareness and greatly improves its ability to accurately identify target regions. Furthermore, compared to lightweight real-time methods, our GMBINet demonstrates superior performance. For instance, GMBINet better preserves structural integrity compared to MINet, which exhibits severe discontinuities in the detection results (highlighted in the blue box in  Fig. \ref{fig:STOA}). This advantage primarily stems from the BPFI within GMBI, which effectively enhances target region perception and expands the network’s receptive field via efficient scale-aware information interaction.

\subsubsection{Rail Surface Defects Detection} 
\begin{figure}   
    \centering    
    \includegraphics[width=0.85\linewidth]{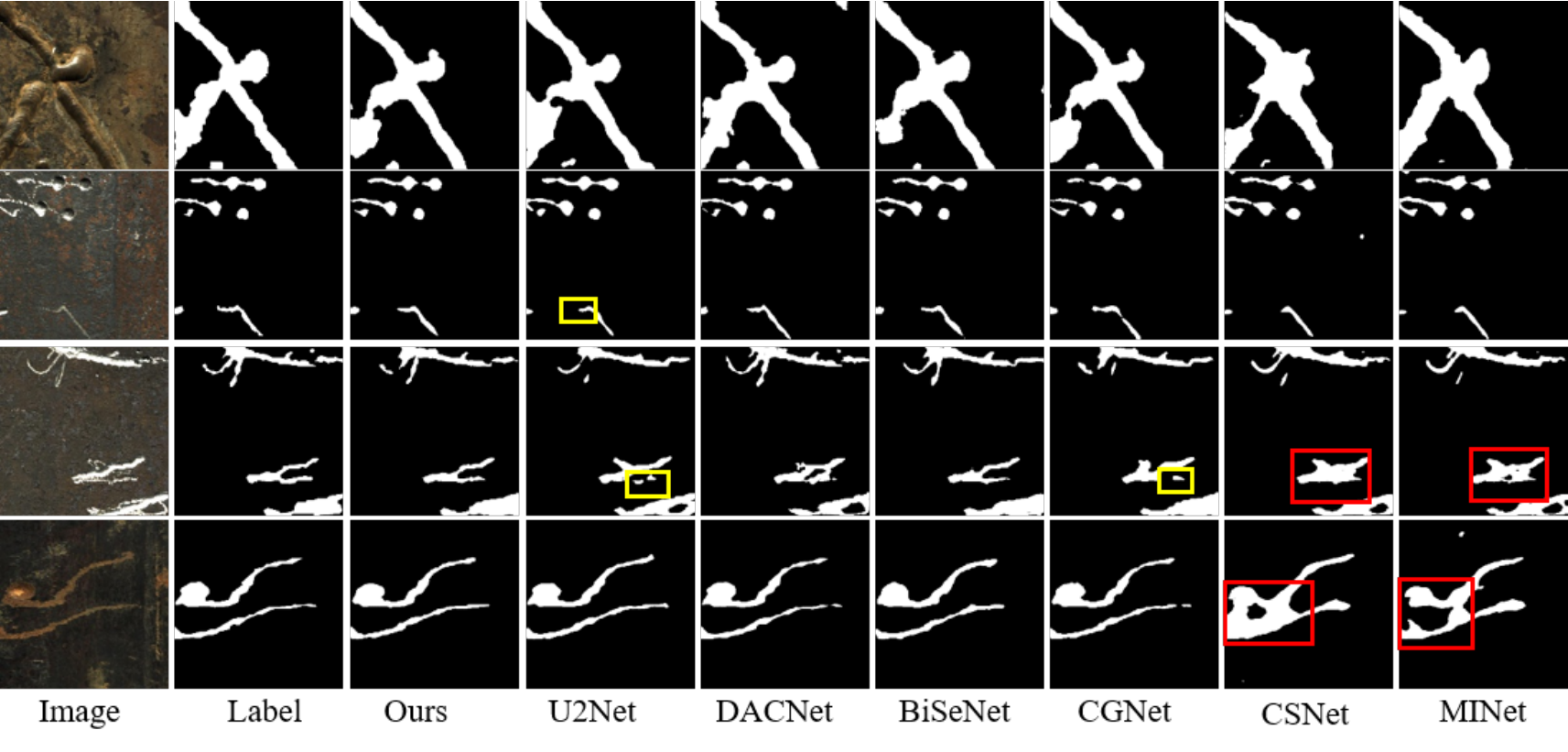}
    \caption{Qualitative results for different methods on the NRSD-MN dataset. The yellow box highlights under-segmentation areas in U2Net and CGNet, and the red box indicates over-segmentation areas in CSNet and MINet.}
    \label{fig:NRSD-MN}
\end{figure}

To evaluate the generalization capability of GMBINet, we compared it against six representative methods on the NRSD-MN dataset for rail surface defect detection. For fairness, all models were trained from scratch. As shown in Table \ref{tab:NRSD-MN}, our proposed method achieved competitive performance, with a MAE of 0.0251 and an IoU of 0.6963. Furthermore, as illustrated in Fig. \ref{fig:NRSD-MN}, our results align more closely with manual delimitations compared to other methods, particularly those designed for lightweight and real-time applications. For example, GMBINet generates more accurate results, while representative lightweight methods (e.g., CSNet and MINet) experience significant over-segmentation (highlighted in the red boxes). This improvement can be attributed to the BPFI mechanism in our GMBI module, which enhances the model's ability to capture more representative features of the target region by promoting feature interaction across different scales.

\begin{table}[!t]
\begin{center}
\caption{Quantitative comparison of our GMBINet with six representative methods on the NRSD-MN dataset. The best three results in each column have been highlighted in red, green, and blue, respectively.}
\label{tab:NRSD-MN}
\begin{tabular*}{\tblwidth}{@{}CCCCCCCCCCCC@{}}
\toprule
 Methods
    & \makecell{Param\\(M)↓}
    & \makecell{FLOPs\\(G)↓}
    & \makecell{Speed\\(FPS)↑}
    & MAE↓
    & WF↑
    & OR↑
    & SM↑
    & PFOM↑
    & IoU↑\\
\midrule

U2Net \cite{qin2020u2}&	44.01&	150.61&	167&\textcolor{blue}{\textbf{0.0261}}&	\textcolor{blue}{\textbf{0.7704}}&	\textcolor{blue}{\textbf{0.6932}}&	\textcolor{blue}{\textbf{0.8436}}&	\textcolor{green}{\textbf{0.5500}}&	\textcolor{blue}{\textbf{0.6956}}\\ 
DACNet \cite{zhou2021dense}&	98.39& 	571.61& 128&\textcolor{red}{\textbf{0.0212}}&	\textcolor{red}{\textbf{0.8039}}&	\textcolor{red}{\textbf{0.7317}}&	\textcolor{red}{\textbf{0.8622}}&	\textcolor{red}{\textbf{0.6162}}&	\textcolor{red}{\textbf{0.7322}}\\ 
BiSeNet \cite{yu2018bisenet}&13.42& 	15.25& 	\textcolor{green}{\textbf{952}}&0.0283&	0.7517&	0.6682&	0.8281&	0.4583&	0.6702\\ 
CGNet \cite{wu2020cgnet}&	0.49& 	3.55& 	639&0.0268&	0.764&	0.6810&	0.8365&	0.4856&	0.6822\\ 
CSNet \cite{gao2020highly}&	\textcolor{red}{\textbf{0.14}}& 	\textcolor{blue}{\textbf{1.84}}& 	\textcolor{blue}{\textbf{885}}&	0.0372&	0.6846&	0.5947&	0.7835&	0.3937&	0.5957\\
MINet \cite{shen2024minet}&	\textcolor{blue}{\textbf{0.28}}& 	\textcolor{green}{\textbf{0.64}}& 	690&	0.0317&	0.7374&	0.6539&	0.8158&	0.4676&	0.6542 \\
\hline
Ours&	\textcolor{green}{\textbf{0.19}}& 	\textcolor{red}{\textbf{0.39}}& 	\textcolor{red}{\textbf{1048}}&	\textcolor{green}{\textbf{0.0251}}&	\textcolor{green}{\textbf{0.7770}}&	\textcolor{green}{\textbf{0.6954}}&	\textcolor{blue}{\textbf{0.8434}}&	\textcolor{blue}{\textbf{0.4938}}&	\textcolor{green}{\textbf{0.6963}}\\ 

\bottomrule
\end{tabular*}
\end{center}
\end{table}

\subsection{Ablation Study}
In this section, we first evaluate the effectiveness of our GMBI-based backbone, followed by a detailed analysis of the contribution of each component within our GMBI module.
\subsubsection{Backbone Comparison} To assess the impact of different backbones, we design six variants of our GMBINet, as shown in Table \ref{tab:Backbones}. Specifically, we replace our GMBI-based real-time backbone with three classical large-scale DL-based backbones (VGG16 \cite{VGG}, ResNet50 \cite{he2016deep}, and ConvNeXt \cite{liu2022convnet}) and three lightweight DL-based backbones (MobileNet \cite{howard2017mobilenets}, GhostNet \cite{han2020ghostnet}, and ShuffleNet \cite{ma2018shufflenet}), respectively. As expected, replacing the lightweight GMBI-based backbone with more complex architectures generally improves detection accuracy due to stronger feature extraction capabilities. However, these gains come at the expense of significantly increased computational complexity and slower inference speed. For instance, the parameters and FLOPs of our method are only 1.34$\%$ and 0.47$\%$ of ResNet50’s, with the SM value only 0.0033 (0.9334 vs 0.9367) lower than ResNet50’s and $3.57 \times$ faster inference speed. This trend is also observed in VGG16 and ConvNext. Moreover, compared to the three classic lightweight backbones, our model achieves better performance across all computational metrics. Meanwhile, in terms of computational efficiency metrics including parameters, FLOPs, and inference speed, our methods are 9.05$\%$, 10.80$\%$, and 1.26 times MobileNet’s. A similar trend is also observed in GhostNet. Although ShuffleNet surpasses our method in inference speed (1175 FPS vs 1048 FPS), its detection performance is significantly lower, with reductions of 24.60$\%$ in MAE, 2.35$\%$ in WF, 4.84$\%$ in OR, 2.19$\%$ in SM, 9.73$\%$ in PFOM, and 7.47$\%$ in IoU. These results demonstrate that the GMBI-based backbone achieves a favorable balance between detection accuracy and computational efficiency, thus better meeting the requirements of real-world deployment.

\begin{table}[!t]
\begin{center}
\caption{Comparisons with different backbones on SD-Saliency-900 dataset.The best three results in each column have been highlighted in red, green, and blue, respectively.}
\label{tab:Backbones}
\begin{tabular*}{\tblwidth}{@{}CCCCCCCCCCCC@{}}
\toprule
 Backbones
    & \makecell{Param\\(M)↓}
    & \makecell{FLOPs\\(G)↓}
    & \makecell{Speed\\(FPS)↑}
    & MAE↓
    & WF↑
    & OR↑
    & SM↑
    & PFOM↑
    & IoU↑\\
\midrule

VGG16 \cite{VGG}&14.87&80.95&187&\textcolor{blue}{\textbf{0.0129}}&\textcolor{green}{\textbf{0.9243}}&\textcolor{green}{\textbf{0.8676}}&\textcolor{green}{\textbf{0.9358}}&\textcolor{blue}{\textbf{0.9031}}&0.8689\\
ResNet50 \cite{he2016deep} &36.41&55.25&229&\textcolor{red}{\textbf{0.0121}}&\textcolor{green}{\textbf{0.9243}}&	\textcolor{red}{\textbf{0.8715}}&\textcolor{red}{\textbf{0.9367}}&\textcolor{red}{\textbf{0.9135}}&\textcolor{red}{\textbf{0.8766}}\\
ConvNeXt \cite{liu2022convnet} &24.53&17.56&276&0.0135&\textcolor{blue}{\textbf{0.9241}}&	0.8627&\textcolor{blue}{\textbf{0.9315}}&0.8593&\textcolor{blue}{\textbf{0.8695}}\\
\hline 
MobileNet \cite{howard2017mobilenets}&2.21&3.52&\textcolor{blue}{\textbf{833}}&0.0139&0.9209&0.8449& 0.9297&0.8805&0.8512\\ 
GhostNet \cite{han2020ghostnet} &\textcolor{blue}{\textbf{1.25}}&\textcolor{blue}{\textbf{1.50}}&817&0.0145&0.9116&0.8373& 0.9196 &0.8600&0.8408\\ 
ShuffleNet \cite{ma2018shufflenet}&\textcolor{green}{\textbf{0.44}}&\textcolor{green}{\textbf{0.82}}&\textcolor{red}{\textbf{1175}}&0.0157&0.9060& 0.8254&0.9134&0.8272&0.8121\\ 
\hline
Ours&\textcolor{red}{\textbf{0.19}}&\textcolor{red}{\textbf{0.39}}&\textcolor{green}{\textbf{1048}}&\textcolor{green}{\textbf{0.0126}}&\textcolor{red}{\textbf{0.9273}}& \textcolor{blue}{\textbf{0.8653}}&0.9334&\textcolor{green}{\textbf{0.9077}}&\textcolor{green}{\textbf{0.8728}}\\

\bottomrule
\end{tabular*}
\end{center}
\end{table}

\subsubsection{Ablation Study for GMBI Module} \textbf{Group Multiscale Feature Extraction Architecture}. To validate the effectiveness of the GMBI module architecture, we design four variants. As illustrated in Fig. \ref{fig: MultiBranch}(c), to demonstrate the effectiveness of the group-based multiscale strategy, we first replace our multiscale feature extraction block with a general 3×3 DWConv operation, denoted as ``w/o MS". Subsequently, we substitute the group-based multiscale strategy with the classical branch-based multiscale strategy, where each branch processes the input tensor directly without the splitting operation, marked as ``w/ branch". As shown in Table \ref{tab:Variants}, both multiscale methods outperformed ``w/ MS", demonstrating the effectiveness of multiscale features in enhancing surface defect detection accuracy. While ``w/ Branch" offers a slight performance improvement over our GMBINet in terms of MAE (0.0130 vs 0.0126) and SM (0.9334 vs 0.9347), it incurs significantly higher computational resources (Param: 0.71 M vs 0.19 M, FLOPs: 1.42 G vs 0.39 G) and results in lower inference speed (623 FPS vs 1048 FPS). Based on this analysis, the group-based multiscale strategy was ultimately adopted in the final GMBI module.

\begin{table}[!t]
\begin{center}
\caption{Comparisons with different variants on the SD-Saliency-900 dataset. The best three results in each column have been highlighted in red, green, and blue, respectively.}
\label{tab:Variants}
\begin{tabular*}{\tblwidth}{@{}CCCCCCCCCCCC@{}}
\toprule
 Methods
    & \makecell{Param\\(M)↓}
    & \makecell{FLOPs\\(G)↓}
    & \makecell{Speed\\(FPS)↑}
    & MAE↓
    & WF↑
    & OR↑
    & SM↑
    & PFOM↑
    & IoU↑\\
\midrule
w/o MS&0.19&0.39&\textcolor{red}{\textbf{1138}}&0.0157&0.9060&0.8275& 0.9149&0.8428&0.8297\\ 
w/ Branch&0.71&1.42&623&\textcolor{green}{\textbf{0.0130}}&\textcolor{green}{\textbf{0.9268}}&\textcolor{blue}{\textbf{0.8641}}& \textcolor{red}{\textbf{0.9347}}&0.8906&\textcolor{green}{\textbf{0.8712}}\\ 
\hline
w/o Interaction &0.19&0.39&\textcolor{green}{\textbf{1098}}&0.0150&0.9095&0.8342&0.9202&0.8534&0.8491\\ 
w/ FG &0.19&0.39&\textcolor{blue}{\textbf{1074}}&0.0135&0.9125&0.8436& 	0.9224&0.8751&0.8629\\ 
w/ BE &0.19&0.39&1085&0.0141&0.9138&0.8395& 	0.9238&0.8813&0.8605\\ 
\hline
w/ Sum &0.19&0.39&1072&0.0133&0.9232&0.8635& 0.9293&\textcolor{blue}{\textbf{0.8971}}&0.8668\\ 
w/ Multiply	&0.19&0.39&1067&0.0135&0.9218& 	0.8622&0.9276&0.8943&0.8587\\ 
w/ Concat&	0.53&0.89&932&\textcolor{blue}{\textbf{0.0132}}&\textcolor{blue}{\textbf{0.9247}}& \textcolor{green}{\textbf{0.8644}}&\textcolor{blue}{\textbf{0.9322}}&\textcolor{green}{\textbf{0.8994}}&\textcolor{green}{\textbf{0.8698}}\\ 
\hline
Ours&0.19&0.39&1048&\textcolor{red}{\textbf{0.0126}}&\textcolor{red}{\textbf{0.9273}}& 	\textcolor{red}{\textbf{0.8653}}&\textcolor{green}{\textbf{0.9334}}&\textcolor{red}{\textbf{0.9077}}&\textcolor{red}{\textbf{0.8728}}\\

\bottomrule
\end{tabular*}
\end{center}
\end{table}

\textbf{Bidirectional Progressive Feature Interactor}. To verify the effectiveness of our BPFI, three variants were introduced: ``w/o interaction", where both forward guidance and backward enhancement are excluded; ``w/ FG", where only forward guidance is employed in the GMBI module; and ``w/ BE", where only backward enhancement is used in the GMBI module. As demonstrated in Table \ref{tab:Variants}, both ``w/ FG" and ``w/ BE" exhibit superior performance compared to ``w/o" interaction", with only a marginal decrease in GPU inference speed. Moreover, we have achieved additional performance gains by concurrently integrating forward guidance and backward enhancement within the GMBI module. This demonstrates that these two information interaction strategies are complementary and effectively facilitate communication between multiscale feature maps. This, in turn, enables the model to better capture the intrinsic features of the target, which aligns with human cognitive understanding, where forward cues and backward feedback collaborate to refine perception.

\textbf{Feature Interaction Types}. To validate the effectiveness of our EWMS operation, we compare it with three representative methods: element-wise summation, element-wise multiplication, and Concat, marked as ``w/ Sum", ``w/ Multiply", and ``w/ Concat". As shown in Table \ref{tab:Variants}, ``w/ Concat" yields better performance than both ``w/ Sum" and ``w/ Multiply". However, ``w/ Concat" incurs high computational costs, with its parameters and FLOPs being 2.65 and 2.38 times those of ``w/ Sum" and ``w/ Multiply", respectively. This is mainly because the ``w/ Concat" operation often requires additional convolutions for feature smoothing and channel adjustment. Moreover, our EWMS operation achieves the best performance, which can be attributed to its efficient facilitation of the interaction between semantic and detailed features. 

\subsubsection{Scale Dimensions Analysis}
The scale dimension $n$ governs the multiscale representation in feature extraction and interaction processes, where a larger $n$ facilitates broader scale integration in multiscale modeling. As shown in Table \ref{tab: SD}, the model maintains a consistent parameter size (0.19 M) and computational complexity (0.39 G) in all $n$ settings.  This is due to the group strategy in the GMBI module, which fixes channel dimensions during multiscale operations, effectively decoupling scale diversity from computational cost. However,  as $n$ increases, the inference speed gradually decreases from 1127 FPS ($n=1$) to 1012 FPS ($n=16$). We attribute this to the increased memory access cost caused by the increased convolution operations, which reduces the cache efficiency \cite{ma2018shufflenet}. Furthermore, detection performance improves initially with increasing $n$, reaching its peak at $n=4$. Beyond this point, performance decreases slightly (e.g., IoU drops from 0.8728 to 0.8687 at $n=16$). This phenomenon can be attributed to excessive channel splitting, where excessive partitioning of feature channels reduces the information capacity within each group, ultimately affecting the model's ability to capture discriminative patterns. Considering both inference speed and detection performance, we select $n=4$ as the optimal value for multiscale representation.
\begin{table}[!t]
\begin{center}
\caption{Quantitative comparison of different scale dimensions settings on the  SD-Saliency-900 dataset.The best three results in each column have been highlighted in red, green, and blue, respectively.}
\label{tab: SD}
\begin{tabular*}{\tblwidth}{@{}CCCCCCCCCCCC@{}}
\toprule
 \makecell{Scale\\dimensions}
    & \makecell{Param\\(M)↓}
    & \makecell{FLOPs\\(G)↓}
    & \makecell{Speed\\(FPS)↑}
    & MAE↓
    & WF↑
    & OR↑
    & SM↑
    & PFOM↑
    & IoU↑\\
\midrule
1&0.19&0.39&\textcolor{red}{\textbf{1127}}&0.0154&0.9051&0.8349&0.9170&0.8839&0.8285\\ 
2&0.19&0.39&\textcolor{green}{\textbf{1093}}&0.0136&0.9177&0.8542&0.9272&\textcolor{blue}{\textbf{0.8951}}&0.8501\\
4&0.19&0.39&\textcolor{blue}{\textbf{1048}}&\textcolor{red}{\textbf{0.0126}}&\textcolor{red}{\textbf{0.9273}}&\textcolor{red}{\textbf{0.8653}}&\textcolor{red}{\textbf{0.9334}}&\textcolor{red}{\textbf{0.9077}}&\textcolor{red}{\textbf{0.8728}}\\
8&0.19&0.39&1024&\textcolor{green}{\textbf{0.0132}}&\textcolor{green}{\textbf{0.9187}}&\textcolor{green}{\textbf{0.8561}}&\textcolor{green}{\textbf{0.9285}}&\textcolor{green}{\textbf{0.8976}}&\textcolor{green}{\textbf{0.8713}}\\
16&0.19&0.39&1012&\textcolor{blue}{\textbf{0.0134}}&\textcolor{blue}{\textbf{0.9180}}&\textcolor{blue}{\textbf{0.8553}}&\textcolor{blue}{\textbf{0.9275}}&0.8941&\textcolor{blue}{\textbf{0.8687}}\\
\bottomrule
\end{tabular*}
\end{center}
\end{table}


\subsection{Practical Implications}

Surface defects are prevalent in industrial manufacturing, where real-time automated detection is critical for enhancing productivity and enabling intelligent control. The primary application of our GMBINet is real-time surface defect detection. Moreover, the core innovation of this work lies in the exploration of lightweight multiscale feature extraction and cost-efficient cross-scale feature interaction. To achieve this, we introduced the GMBI module. The GMBI module presents a structurally simple design (i.e., single input and single output), allowing it to function as a plug-and-play component that can be seamlessly integrated into existing network architectures. Its lightweight nature and effective multiscale interaction make it suitable for a variety of industrial vision tasks beyond surface defect detection, such as defect classification, object tracking, and semantic segmentation. For example, we construct a GMBI-based backbone by replacing the standard residual blocks in ResNet50 with GMBI modules for defect classification. As shown in Table \ref{tab: classification}, the GMBI-based network achieves a substantial acceleration in inference speed (1817 FPS vs 357 FPS) and a dramatic reduction in both parameter count (0.19 M vs 23.52 M) and computational cost (0.31 G vs 21.59 G), with only a minor drop in classification accuracy (0.9806 vs 0.9944). This optimal trade-off between performance and efficiency highlights the GMBI-based backbone’s potential for deployment in real-time, resource-constrained industrial environments.

\begin{table}[!t]
\begin{center}
\caption{Surface Defects Classification on the NEU-CLS Dataset. The best three results in each column have been highlighted in red, green, and blue, respectively.}
\label{tab: classification}
\begin{tabular*}{\tblwidth}{@{}CCCCCCCCCCCC@{}}
\toprule
    Networks
    & \makecell{Param\\(M)↓}
    & \makecell{FLOPs\\(G)↓}
    & \makecell{Speed\\(FPS)↑}

    & Accuracy↑
    & Precision↑
    & Recall↑
    & F1-Score↑\\
\midrule
VGG16 \cite{VGG}&138.36&80.30&262&0.9861&0.9862&0.9861&0.9861\\
ResNet50 \cite{he2016deep}&23.52&21.59&357&\textcolor{red}{\textbf{0.9944}}&\textcolor{red}{\textbf{0.9945}}&\textcolor{red}{\textbf{0.9944}}&\textcolor{red}{\textbf{0.9944}}\\
ConvNeXt \cite{liu2022convnet}&27.80&23.27&346&\textcolor{green}{\textbf{0.9889}}&\textcolor{green}{\textbf{0.9892}}&\textcolor{green}{\textbf{0.9889}}&\textcolor{green}{\textbf{0.9889}}\\
Swin-T \cite{liu2021swin}&15.56&18.86&391&\textcolor{blue}{\textbf{0.9889}}&\textcolor{blue}{\textbf{0.9891}}&\textcolor{blue}{\textbf{0.9889}}&\textcolor{blue}{\textbf{0.9889}}\\
\hline
MobileNet \cite{howard2017mobilenets}&\textcolor{blue}{\textbf{2.23}}&1.70&1087&0.9778&0.9795&0.9778&0.978\\
GhostNet \cite{han2020ghostnet}&5.18&\textcolor{blue}{\textbf{0.80}}&\textcolor{blue}{\textbf{1526}}&0.9667&0.9679&0.9667&0.9666\\
ShuffleNet \cite{ma2018shufflenet}&\textcolor{green}{\textbf{0.35}}&\textcolor{red}{\textbf{0.23}}&\textcolor{red}{\textbf{2239}}&0.9472&0.9484&0.9472&0.9473\\
\hline
Ours&\textcolor{red}{\textbf{0.19}}&\textcolor{green}{\textbf{0.31}}&\textcolor{green}{\textbf{1817}}&0.9806&0.9816&0.9806&0.9806\\
\bottomrule
\end{tabular*}
\end{center}
\end{table}



\section{Conclusion}
\label{sec:conclusion}
This paper presents GMBINet, a lightweight encoder–decoder network for real-time steel surface defect detection. At its core is the GMBI module, which leverages a hierarchical group-wise design to extract scale-aware feature representations without increasing computational complexity. To further enhance cross-scale information flow, the GMBI module integrates a BPFI mechanism with a parameter-free EWMS operation, enabling effective feature interaction with minimal overhead. Further evaluation on the NEU-CLS classification dataset confirms its versatility and potential for broader industrial vision tasks. These results highlight its suitability for deployment in real-time and resource-constrained industrial environments.

To further enhance the applicability of the proposed method under challenging conditions (e.g., poor illumination, reflective surfaces, depth ambiguity), future work will focus on improving model robustness through multimodal data fusion by integrating complementary signals such as depth and thermal imaging. Additionally, we plan to explore hardware-aware optimization techniques (e.g., pruning, knowledge distillation, quantization) to facilitate efficient deployment on edge devices within smart manufacturing systems.
\section*{Acknowledgements}
This work was supported in part by the National Natural Science Foundation of China 62376042, the Natural Science Foundation of Chongqing CSTB2024NSCQ-KJFZZDX0036, the Key Special Project for Technological Innovation and Application Development in Chongqing CSTB2023TIAD-KPX0050, CSTB2022TIAD-KPX0176.

\bibliographystyle{unsrt}

\bibliography{cas-refs}



\end{document}